\documentclass[runningheads]{llncs}
\usepackage[T1]{fontenc}

\usepackage{url}
\usepackage{cite}
\usepackage{amsmath,amssymb,amsfonts}
\usepackage{algorithmic}
\usepackage{graphicx}
\usepackage{textcomp}
\usepackage{xcolor}
\usepackage{algorithm}
\usepackage{xcolor}
\usepackage{soul}
\usepackage{amsmath}
\usepackage{enumitem}
\usepackage{array}
\usepackage{multirow}
\usepackage{booktabs}
\usepackage{array}

\begin{document}
\mainmatter              
\title{Understanding Fairness-Accuracy Trade-offs in Machine Learning Models: Does Promoting Fairness Undermine Performance?}
%
\titlerunning{Understanding Fairness-Accuracy Trade-offs in Machine Learning Models}  
%
\author{Junhua Liu\inst{1}\orcidID{0000-0003-4477-7439}
\and\\ Roy Ka-Wei Lee\inst{2}\orcidID{0000-0002-1986-7750}
\and\\ Kwan Hui Lim\inst{2}\orcidID{0000-0002-4569-0901}}

\authorrunning{J. Liu et al.}  

\institute{
Forth AI\\
\and
Singapore University of Technology and Design\\
\email{j@forth.ai, \{roy\_lee, kwanhui\_lim\}@sutd.edu.sg}}
 
\maketitle

\begin{abstract}

Fairness in both Machine Learning (ML) predictions and human decision-making is essential, yet both are susceptible to different forms of bias, such as algorithmic and data-driven in ML, and cognitive or subjective in humans. In this study, we examine fairness using a real-world university admissions dataset comprising 870 applicant profiles, leveraging three ML models: XGB, Bi-LSTM, and KNN, alongside BERT embeddings for textual features. To evaluate individual fairness, we introduce a consistency metric that quantifies agreement in decisions among ML models and human experts with diverse backgrounds. Our analysis reveals that ML models surpass human evaluators in fairness consistency by margins ranging from 14.08\% to 18.79\%. 
Our findings highlight the potential of using ML to enhance fairness in admissions while maintaining high accuracy, advocating a hybrid approach combining human judgement and ML models.

\end{abstract}

%

\section{Introduction}
Fairness has emerged as a critical concern in terms of both the predictive outputs of Machine Learning (ML) models and the decisions made by human experts~\cite{wang2023survey,li2023fairness,caton2024fairness,mehrabi2021survey}. In the context of ML models, challenges in fairness often stem from data biases and algorithmic limitations, which can propagate or exacerbate existing inequities. Conversely, fairness in human decision-making is inherently influenced by subjective judgment and cognitive biases, making it vulnerable to inconsistency and error. These two sources of unfairness necessitate rigorous investigation into both computational and human-centric frameworks to mitigate bias and ensure equitable outcomes.

To investigate fairness, this work utilizes a real-world university admission dataset with 870 unique profiles and three effective ML models, namely, Extreme Gradient Boosting (XGB), Bi-directional Long Short-Term Memory (Bi-LSTM) and K-Nearest Neighbours (KNN). The textual features of the profiles are encoded into high-dimensional embeddings using the Bidirectional Encoder Representations from Transformers (BERT), enabling the extraction of rich contextual representations for downstream predictive tasks. In our case, we focus on the important but challenging task of admission offer decision making.

\begin{figure*}[t]
    \centering
    \includegraphics[width=\linewidth]{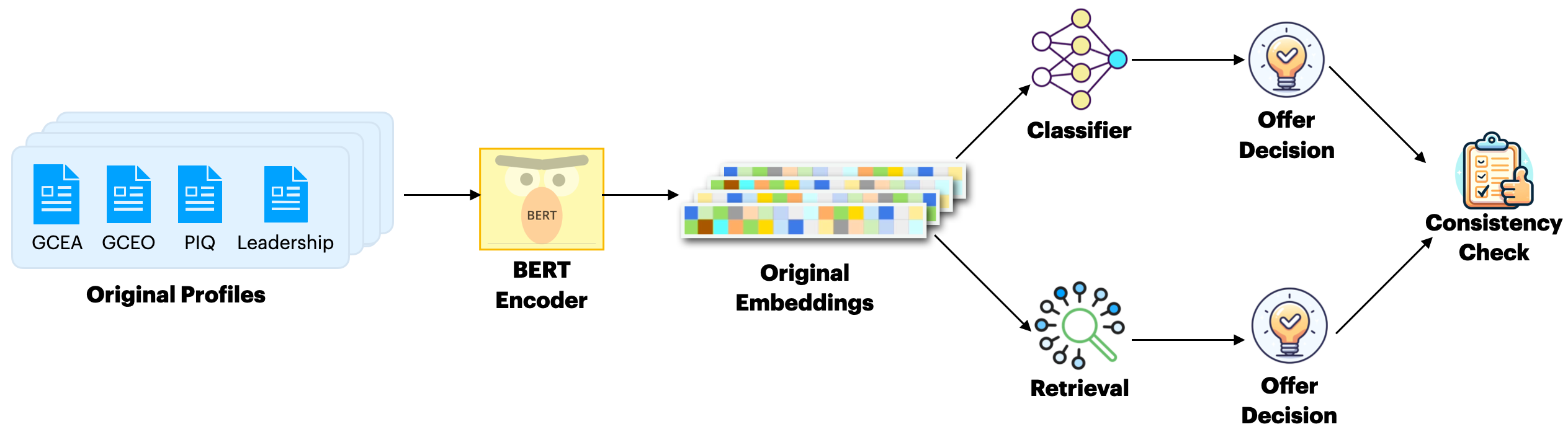}
    \caption{Proposed pipelines to assess Individual Fairness}
    \label{fig:pipelines}
\end{figure*}

We examine individual fairness in university admission offer decisions for a heterogeneous group of applicants, characterized by diverse backgrounds, leadership qualities, and academic achievements. Using a standardized set of features for comparison, we assess individual fairness by measuring decision consistency through a consistency score~\cite{pmlr-v28-zemel13}, which quantifies how similar applicants with comparable profiles are treated by human experts or ML models. Our results reveal that both XGBoost and Bi-LSTM outperform human experts significantly, achieving higher consistency scores by a margin of 14.08\% to 18.79\%. These findings suggest that ML models demonstrate superior reliability in ensuring fair treatment of similar applications.

\section{Related Work}

\textbf{Algorithmic Fairness}. Individual fairness emphasizes treating similar individuals similarly and is closely related to differential privacy through the Lipschitz condition~\cite{10.1145/2090236.2090255,dwork06}. Fairness definitions often vary depending on the worldview and can typically be derived from confusion matrices~\cite{aifairness}. The increasing use of large language models (LLMs) in decision-making raises concerns about potential human-like biases these models may adopt~\cite{science}, necessitating investigation and effective mitigation strategies. Various methods to address biases have been explored, including robust data preprocessing~\cite{debiasing2023education}, fair representation learning~\cite{pmlr-v28-zemel13}, and auditing human decision processes~\cite{alur2023auditing}.

\textbf{Fairness in University Admissions}. Previous work on fairness in admissions often focuses on qualitative analyses~\cite{zimdars2010fairness}, general policy evaluations~\cite{pitman2016understanding}, or assessments of specific factors like standardized testing~\cite{van2023analysis}. However, these studies tend to involve limited user groups or geographic contexts, thus restricting their generalizability. Other studies compare perceptions of fairness in AI-driven versus human-driven admissions decisions~\cite{marcinkowski2020implications}. Recent methods include Puranik et al.’s Fair-Greedy policy~\cite{puranik2022dynamic}, balancing score maximization with representation of under-represented groups, and Bhattacharya et al.’s empirical evaluation of socioeconomic biases, identifying disparities in admission thresholds among demographic groups~\cite{bhattacharya2017university}.

\section{Data Processing}

\subsection{Dataset}
We use the university admission dataset introduced in~\cite{liu-ASONAM25a}, which consists of applicant profiles from a specific admission cycle year. 
The dataset includes a total of 870 unique profiles that come with four features: General Certificate of Education Advanced Level (GCEA), General Certificate of Education Ordinary Level (GCEO), Personal Insight Questions (PIQ) and Leadership Experience. We concatenate all content and form a \texttt{Combined} document. There is also the label Type, which is either Offered or Not Offered. The dataset is then split into  80\% training, 10\% validation, and 10\% test for experimentation purposes.

\subsection{Feature Embedding}
To prepare the data for fairness assessment, we generate contextual embeddings using BERT (bert-base-uncased). For each feature, we generated BERT embeddings using the following process: the text was tokenized and padded to a maximum length of 512 tokens, then processed through BERT to obtain contextual embeddings. We extracted the [CLS] token embedding (768-dimensional vector) as the representation for each feature. The individual embeddings were then concatenated to create a combined representation (3840-dimensional vector) for each profile, preserving the contextual information. 
The final output included individual embeddings for each feature and embeddings for the concatenated feature. 

\subsection{Human Decisions}
We use three human decision points, such as Shortlisting (SL), Admission Recommendation (AR) and Offer (OF), as targets to assess prediction accuracy and consistency. Specifically, we mapped the categorical labels to binary values, where "Shortlisted", "Recommended" and "Offered" correspond to 1, and 0 otherwise, respectively.

\section{Individual Fairness}
\label{sec:if}

\subsection{Formulation}
Individual fairness is grounded in the principle that similar individuals should receive similar treatment in decision-making processes. Following~\cite{pmlr-v28-zemel13}, we operationalize this concept through a consistency metric that evaluates how similarly a classifier treats applicants with comparable profiles. Figure~\ref{fig:pipelines} illustrates our proposed pipeline for assessing individual fairness.

Formally, given a set of N applicants, we define the consistency score $\mathcal{C}$ of a classifier as:

\begin{equation}
    \mathcal{C} = 1 - \frac{1}{N} \sum_{i=1}^N |\hat{y}_i - \frac{1}{k} \sum_{j \in knn(i)} \hat{y}_j|
\end{equation}

where $\hat{y}_i$ represents the classifier's prediction for applicant $i$, and $\frac{1}{k} \sum_{j \in knn(i)} \hat{y}_j$ computes the average of predictions among the k-nearest neighbors (i.e. most similar profiles), of applicant $i$. This formulation yields a score between 0 and 1, where higher values indicate greater consistency in treating similar cases similarly. The scores effectively reflect the fairness consistency by measuring mean absolute differences between a classifier's predictions to the target profiles and the average predictions made for most similar profiles.

\subsection{Research Questions}
The rest of this section discusses our investigation on the following Research Questions (RQ):

\begin{itemize}
    \item \textbf{RQ1:} Can Cognitive Bias in human decisions be systematically identified?
    \item \textbf{RQ2:} Are ML models more consistent in making decisions than human?
    \item \textbf{RQ3:} Can ML models outperform human in making admission decisions?
\end{itemize}




\begin{table}[t]
\caption{Individual Fairness: classification performance and fairness consistency. Best results are bolted. Abbreviations: P=precision; R=Recall; A=Accuracy; C=Consistency}

\begin{center}
\scalebox{1}{
\begin{tabular}{ccccccc}
\toprule
\textbf{Model} & \textbf{P} & \textbf{R} & \textbf{F1} & \textbf{A} 
& \textbf{C(AR)} & \textbf{C(OF)} \\ \toprule
\multicolumn{7}{c}{Human Decisions}\\ \midrule
Shortlist(SL) & \textbf{0.8464} & \textbf{0.8418} & 0.8156 & 0.8155 & - & - \\ 
Adm. Rec. (AR) & 0.8011 & 0.8193 & 0.8321 & 0.8087 & 0.5632 & -\\ 
Offer (OF) & - & - & - & - & - & 0.6023 \\ \midrule
\multicolumn{7}{c}{ML Models}\\ \midrule
KNN (k=5) & 0.6886 & 0.6707 & 0.6897 & 0.6897 & - & - \\
XGB & 0.7902 & 0.7859 & 0.7878 & 0.7931 & 0.7521 & 0.7477 \\ 
Bi-LSTM & 0.8291 & 0.8178 & \textbf{0.8176} & \textbf{0.8276} & \textbf{0.8073} & \textbf{0.7797} \\ 
\bottomrule
\end{tabular}}
\end{center}
\label{table:experiment}
\end{table}

\subsection{Methodology}
Our evaluation framework examines individual fairness across the entire admission pipeline by analyzing three human decision points, such as shortlisting (SL), admission recommendations (AR), and final offers (OF), alongside two model-based classifiers (XGB and BiLSTM) and a retrieval-based k Nearest Neighbor (KNN) classifier. The foundation of our methodology lies in the construction of a feature-reranked similarity matrix, which provides a robust measure of application similarity based on embedded features. This similarity matrix enables the identification of nearest neighbors for consistency evaluation.

\subsubsection{Implementation:}
For model training and evaluation, we choose three different approaches, namely, traditional classifiers, neural networks and retrieval-based classifiers. For each category, we choose an empirically effective model from each category, such as XGB, BiLSTM, and KNN. 
The XGB classifier is implemented using the xgboost library. The BiLSTM model was implemented using PyTorch, featuring a bidirectional LSTM layer followed by linear layers with ReLU activation and LogSoftmax output. For the KNN approach, we utilized the FAISS library for similarity search with in-built memory-efficient processing predictions in batches.

\subsubsection{Hyperparameters:} 
We use the validation set to perform hyperparameter search with randomised scheme. Each run takes 20 epochs with a patience of 5 (i.e. 5 epochs without improvement). Best sets are selected using average accuracy. 

\subsubsection{Metrics:}
For each decision point and model, we compute both traditional classification metrics (precision, recall, F1-score, and accuracy) and consistency score. The consistency evaluation uses k=5 nearest neighbors, chosen to balance between local similarity and statistical stability. We maintain consistent feature representations across all models by using the same embedded feature space, ensuring fair comparisons between human decisions and machine learning predictions. This approach allows us to systematically assess how different decision-making approaches align with the principle of treating similar applications similarly, while also measuring their overall classification performance.

\subsection{Experimental Results}
Table~\ref{table:experiment} summarizes the experimental results for classification performance and individual fairness consistency.

\subsubsection{Human Decisions:}
The shortlisting decisions (SL) made by human evaluators achieve the highest precision among all decision points and models, with a precision of 84.64\%. This high precision indicates strong reliability in correctly identifying qualified candidates at the initial stage of the admission process. The admission recommendation (AR) stage exhibits slightly lower but still robust performance, with an accuracy of 80.87\% and an F1-score of 83.21\%. These metrics suggest consistent and effective decision-making during the early phases of candidate evaluation process.

\subsubsection{Machine Learning Models:} 
Among the ML models, Bi-LSTM demonstrates superior performance, achieving an F1-score of 81.76\% and an accuracy of 82.76\%. These results closely match those of human decisions, with Bi-LSTM attaining a comparable F1-score and even surpassing human decisions in accuracy. This indicates that the Bi-LSTM model is effective in replicating human-like decision-making patterns while offering enhanced accuracy.

In contrast, the KNN model with $k=5$ shows significantly lower performance, with both F1-score and accuracy at 68.97\%. This suggests that a non-parametric retrieval model like KNN is less capable of capturing the complex patterns inherent in human decision-making, highlighting the importance of model selection in predictive tasks.

\subsubsection{Fairness Consistency:} 

In terms of fairness consistency, we observe notable variations between human decisions and machine learning model predictions. The consistency score for the admission recommendations (AR) stage is 56.32\%, while the final offer decisions (OF) exhibit a slightly higher consistency at 60.23\%. These figures suggest that human decision-making maintains moderate levels of consistency, but there remains considerable room for improvement in ensuring that similar candidates receive similar outcomes.

Conversely, both machine learning models demonstrate substantially higher consistency scores compared to human decisions. The XGB model achieves consistency scores of 75.21\% for AR and 74.77\% for OF stages, indicating a significant improvement over human decision-making in terms of fairness consistency. The Bi-LSTM model performs even better, attaining the highest consistency scores of 80.73\% for AR and 77.97\% for OF. This highlights the potential of machine learning models to provide more consistent and fair evaluations, reducing variability in decision outcomes for similar applicants.

\subsection{Analysis and Interpretations}

\subsubsection{RQ1: Identifying Cognitive bias:}
While early stages of human decisions, such as Shortlisting (SL) and Admission Recommendations (AR) demonstrate strong classification performance i.e. over 8\% in F1-Score and accuracy, their consistency scores are notably lower, with AR's being 56.32\% and OF's being 60.23\%. This disparity between performance metrics and consistency scores suggests potential cognitive biases in the evaluation process. Furthermore, the gradual decrease in consistency from OF to AR stages, i.e. from 60.23\% to 56.32\%, indicates that biases might become more pronounced in later stages of the admission process, possibly due to increased complexity in decision criteria while maintaining high consistency among different review panels.

\subsubsection{RQ2: Consistency comparison between ML and Human:}
The consistency analysis reveals a substantial gap between human and machine learning approaches. Both ML models significantly outperform human decisions in terms of consistency, with Bi-LSTM achieving scores of 80.73\% and 77.97\% for AR and OF respectively, compared to human scores of 56.32\% (AR) and 60.23\% (OF). This represents an improvement of approximately 24\% in consistency for AR decisions and 18\% for OF decisions. The XGB model also demonstrates superior consistency with 75.21\% for AR and 74.77\% for OF, suggesting that machine learning approaches inherently provide more standardized evaluation patterns. This substantial difference in consistency scores highlights the potential role of ML models in reducing decision variability.

\subsubsection{RQ3: Accuracy comparison between ML and Human:}
The performance comparison between human decisions and ML models yields nuanced insights. While human shortlisting achieves the highest precision of 84.64\%, the Bi-LSTM model demonstrates comparable overall performance with the highest accuracy of 82.76\% and F1-score of 81.76\%. Furthermore, the Bi-LSTM's performance metrics closely match or exceed human decision points while maintaining significantly higher consistency scores. The XGB model, while showing lower performance metrics with an F1-score of 78.78\%, still maintains higher consistency than human decisions, suggesting a potential trade-off between performance and consistency that varies across different ML architectures.


In summary, the experimental results demonstrate that while human decisions maintain high classification performance, they show lower consistency compared to ML approaches. These findings suggest that ML models could serve as valuable decision support tools, particularly in maintaining consistency across the admission process while preserving high accuracy standards. The results advocate for a carefully designed hybrid approach that combines human expertise with ML-driven consistency checks to optimize both performance and fairness in admission decisions.







\section{Conclusion}

Fairness in decision-making is critical, whether conducted by artificial intelligence or human. Our work investigates individual fairness, using a consistency score to evaluate decision-making among a diverse group of subjects. Experimental results reveals that ML models such as XGBoost and BiLSTM achieve significantly higher consistency scores than human experts. This finding suggests that ML models can provide more consistent treatment of similar applications, addressing any potential subjectivity and cognitive biases inherent in human decision-making.


Our findings have practical implications for the design of admission systems, suggesting that ML models could serve as valuable decision support tools, particularly in initial screening stages where consistency is crucial. The models could help identify potential inconsistencies in human decisions, prompting additional review in cases where machine and human assessments diverge. Furthermore, the quantification of consistency scores provides a new dimension for evaluating and improving admission processes, beyond traditional accuracy metrics.



\vspace{3mm}
{
\small
{\noindent\bf Acknowledgments.} 
This research is supported in part by the Ministry of Education, Singapore (MOE), under its Academic Research Fund Tier 2 (Award No. MOE-T2EP20123-0015), and the Singapore University of Technology and Design (SUTD) under grant RS-MEFAI-00011. Any opinions, findings and conclusions, or recommendations expressed in this material are those of the authors and do not reflect the views of MOE or SUTD.
}

%
%
\bibliographystyle{splncs04}
\bibliography{refShort}








\end{document}